\documentclass[11pt,conference]{IEEEtran}
\usepackage{amsmath}
\usepackage{amssymb}
\usepackage{booktabs}
\usepackage{array}
\usepackage{url}
\usepackage{cite}
\usepackage{microtype}
\usepackage{graphicx}
\usepackage{xcolor}
\usepackage{tikz}
\usepackage{pgfplots}
\usepackage{multirow}
\usepackage{tabularx}
\usepackage{caption}

\usetikzlibrary{shapes.geometric, arrows.meta, positioning, fit, backgrounds, calc, patterns}
\pgfplotsset{compat=1.18}

\begin{document}
	
	\title{An Affordable AI-Integrated Smart Cane for Multimodal Mobility Assistance of Visually Impaired Users}

\author{
	\IEEEauthorblockN{
		Ali Akarma\IEEEauthorrefmark{1}\IEEEauthorrefmark{2},
		Adeel Ahmad\IEEEauthorrefmark{1},
		Toqeer Ali Syed\IEEEauthorrefmark{1}
	}
	\IEEEauthorblockA{\IEEEauthorrefmark{1}
		AI Center, Faculty of Computer and Information Systems, \\
		Islamic University of Madinah, Madinah 42351, Saudi Arabia\\
		Emails: 443059463@stu.iu.edu.sa, 443057803@stu.iu.edu.sa,
		toqeer@iu.edu.sa
	}
	\IEEEauthorblockA{\IEEEauthorrefmark{2}
		AI V\&V Lab, King Fahd University of Petroleum and Minerals,
		Dhahran 31261, Saudi Arabia
	}
}

	\maketitle
	
	\begin{abstract}
		Visual impairment affects over 2.2 billion people worldwide, yet conventional white canes cannot detect elevated hazards or provide semantic environmental context. Existing AI-assisted navigation systems typically rely on expensive hardware or cloud connectivity, limiting accessibility in resource-constrained settings. This paper presents an affordable (\$88 USD), fully offline AI-integrated smart cane designed for multimodal mobility assistance on an ultra-low-power Raspberry Pi Zero 2W. The system fuses RGB vision sensing with Time-of-Flight (ToF) distance estimation, pairing an INT8-quantized SSD MobileNet V1 model with distance-aware vibrotactile feedback and real-time audio alerts. To ensure operational robustness on constrained hardware, a multiprocessing architecture isolates sensor acquisition, neural inference, and haptic feedback into independent processes with fail-safe sensing support. Experimental evaluation across indoor mobility scenarios demonstrates a macro-averaged F1-score of 0.82 (precision: 0.85, recall: 0.81), a mean end-to-end latency of 330\,ms, and a peak power draw of 2.8\,W. A preliminary usability study with 12 participants (SUS: 78.5, NASA-TLX) demonstrated positive user perception and enhanced obstacle awareness. The proposed prototype validates the feasibility of deploying privacy-preserving, edge-native assistive intelligence for cost-sensitive mobility assistance.
	\end{abstract}
	

	\section{Introduction}
	
	According to the World Health Organization (WHO), more than 2.2 billion people worldwide experience some form of visual impairment, with a substantial proportion facing severe limitations in independent mobility and environmental awareness~\cite{who2023}. For individuals who are blind or have significant vision loss, navigation through unfamiliar indoor and outdoor environments involves complex spatial cognition and wayfinding processes~\cite{loomis2001navigation}, and is frequently accompanied by persistent mobility anxiety, elevated collision risks, and reduced social independence. Beyond physical accessibility, safe mobility directly influences social participation, psychological well-being, educational access, and overall quality of life. This aligns with recent efforts to deploy assistive artificial intelligence and agentic frameworks designed to empower individuals with disabilities and neurodivergence~\cite{Siddiqui2026ADAPT}.
	
	The traditional white cane continues to serve as the most widely adopted assistive mobility tool due to its affordability, portability, and reliability. However, its functionality is inherently limited to tactile interaction with ground-level obstacles. Conventional canes cannot reliably detect elevated or protruding hazards such as hanging signage, partially opened cabinets, tree branches, or obstacles outside the immediate contact region. Furthermore, traditional mobility aids provide minimal semantic awareness regarding the surrounding environment, requiring users to rely heavily on spatial memory, auditory cues, and physical exploration to interpret environmental context. These limitations can increase cognitive workload during navigation, particularly in unfamiliar or densely populated environments.
	
	Recent advances in embedded artificial intelligence and edge computing have motivated the development of intelligent mobility assistance systems capable of providing enhanced environmental awareness. Existing smart navigation solutions have explored ultrasonic sensing, wearable cameras, smartphone-based vision systems, and cloud-assisted object recognition frameworks. While these approaches demonstrate promising assistive capabilities, several practical deployment challenges remain unresolved. Commercial systems such as OrCam MyEye provide advanced scene interpretation functionality but remain financially inaccessible for many users, particularly in low-resource regions~\cite{Lavric2024}. In parallel, many research-oriented smart cane platforms rely on computationally intensive embedded hardware, continuous internet connectivity, or cloud-based inference pipelines. Such architectures introduce increased energy consumption, unpredictable communication latency, privacy concerns, and reduced operational reliability in environments with limited network availability.
	
	Consequently, a significant research gap persists in the development of affordable, reliable, and fully edge-native assistive intelligence systems capable of operating on ultra-low-cost embedded hardware while maintaining practical mobility assistance functionality. In particular, existing studies frequently prioritize object detection performance while placing comparatively less emphasis on reliability-oriented deployment, fail-safe operation, and human-centered assistive feedback under severe hardware constraints.
	
	To address these challenges, this paper presents an affordable AI-integrated smart cane architecture designed for multimodal mobility assistance in visually impaired navigation scenarios. The proposed system combines RGB vision sensing with Time-of-Flight (ToF) distance estimation to provide localized environmental awareness through embedded edge inference, vibrotactile feedback, and audio-based obstacle notification. Unlike cloud-dependent assistive frameworks, the proposed architecture performs all inference locally on a Raspberry Pi Zero 2W platform using an optimized INT8 SSD MobileNet V1 pipeline. In addition, a reliability-oriented multiprocessing framework is introduced to isolate sensing, inference, and feedback subsystems, enabling continued assistive operation during inference slowdowns or partial subsystem failures.
	
	The primary contributions of this work are summarized as follows:
	
	\begin{enumerate}
		\item \textbf{Affordable AI-Integrated Smart Cane Architecture:} We develop a low-cost multimodal mobility assistance system operating entirely on ultra-low-power embedded hardware, achieving a total prototype cost of approximately \$88 USD while maintaining fully offline functionality.
		
		\item \textbf{Multimodal Environmental Awareness Framework:} We integrate embedded computer vision and ToF-based spatial sensing to provide semantic obstacle awareness through synchronized vibrotactile and audio feedback mechanisms with a formally defined proximity-to-haptic encoding scheme.
		
		\item \textbf{Reliability-Oriented Embedded Deployment:} We propose an asynchronous multiprocessing architecture that separates sensing, inference, and feedback pipelines to improve operational stability and maintain fail-safe assistive functionality under constrained hardware conditions.
		
		\item \textbf{Comprehensive Mobility Assistance Evaluation:} We conduct technical, mobility-oriented, and usability-focused evaluations across indoor navigation scenarios, including latency analysis, obstacle detection performance, post-quantization accuracy validation, power profiling, and preliminary user-centered assessment using standardized usability metrics with reported confidence intervals.
	\end{enumerate}

	\section{Related Work}
	\label{sec:related}
	
	Research on intelligent assistive mobility systems for visually impaired individuals has evolved considerably over the past decade, spanning ultrasonic sensing, smartphone-assisted navigation, cloud-based scene understanding, and embedded edge-AI frameworks~\cite{messaoudi2022review,ABIDI2024,Nikanfar2025}. Existing approaches differ substantially in terms of sensing modality, computational architecture, deployment cost, power efficiency, and practical usability in real-world mobility scenarios.
	
	\subsection{Ultrasonic and Range-Sensing Smart Canes}
	
	Early smart cane systems primarily relied on ultrasonic or infrared distance sensing to augment the traditional white cane with basic obstacle awareness capabilities~\cite{shoval1998navbelt}. These systems typically employ lightweight microcontroller platforms and provide low-latency obstacle alerts with minimal energy consumption. Due to their simplicity and affordability, ultrasonic-based mobility aids remain attractive for low-resource deployment scenarios.
	
	Despite these advantages, purely range-based systems provide limited contextual understanding of the surrounding environment. Most ultrasonic frameworks can detect obstacle proximity but cannot distinguish between object categories, identify navigational landmarks, or provide semantic environmental interpretation. Consequently, users may receive frequent proximity alerts without sufficient contextual information to support higher-level navigation decisions, potentially increasing cognitive workload in crowded or unfamiliar environments.
	
	\subsection{Smartphone-Based Assistive Navigation Systems}
	
	The widespread availability of high-performance smartphones has motivated the development of mobile assistive navigation applications using RGB cameras, LiDAR sensors, inertial sensing, and on-device machine learning~\cite{tapu2020survey}. Smartphone-assisted systems benefit from integrated hardware ecosystems and access to increasingly powerful mobile AI accelerators. Several studies have demonstrated real-time object detection, scene understanding, and voice-guided navigation using consumer mobile devices.
	
	However, practical deployment challenges remain significant. Continuous camera-based processing can rapidly increase thermal load and battery consumption during prolonged navigation tasks. In addition, smartphone-based assistive frameworks often require handheld operation or wearable mounting configurations, which may reduce ergonomic comfort and introduce usability concerns during long-term mobility assistance. Device cost and platform fragmentation further complicate accessibility in economically constrained regions.
	
	\subsection{Cloud-Assisted Assistive Intelligence}
	
	To overcome local hardware limitations, some assistive mobility systems employ cloud-assisted inference architectures in which visual or sensor data are transmitted to remote servers for semantic analysis~\cite{chen2019edge}. Cloud-based frameworks enable the use of computationally intensive deep learning models for scene segmentation, object recognition, and route planning while reducing local processing requirements on embedded devices.
	
	Although these approaches can improve inference capability, dependence on persistent network connectivity introduces operational limitations for safety-critical mobility applications. Variable communication latency, bandwidth constraints, intermittent connectivity, and privacy concerns may reduce reliability in crowded urban environments, underground transportation systems, or rural deployment scenarios. As a result, fully cloud-dependent mobility assistance systems may struggle to provide consistent low-latency assistive feedback during real-world navigation.
	
	\subsection{Edge-AI Smart Cane Architectures}
	
	Recent research has increasingly focused on edge-native assistive intelligence systems capable of performing semantic inference directly on embedded hardware platforms~\cite{upadhyaya2024real,technologies14040215}. Smart cane architectures utilizing SSD, YOLO, and lightweight convolutional neural networks on Raspberry Pi and NVIDIA Jetson devices have demonstrated improved environmental awareness and localized obstacle recognition capabilities. Edge-AI systems offer important advantages including offline functionality, improved privacy preservation, and reduced communication latency. These attributes are especially critical in assistive contexts, where privacy-preserving federated and agentic intelligence architectures have been shown to foster disability-inclusive participation and data confidentiality in intelligent urban environments~\cite{syed2026fedagent}. The emerging TinyML paradigm further extends these capabilities by targeting extreme resource constraints characteristic of ultra-low-power embedded devices~\cite{technologies14040215}.
	
	Nevertheless, practical tradeoffs remain between computational performance, energy efficiency, portability, and deployment cost. High-performance embedded platforms such as Jetson Nano or Raspberry Pi 4 can support more advanced neural inference pipelines but often require larger battery systems, active thermal management, and increased physical footprint. Conversely, ultra-low-cost platforms frequently experience constrained memory availability, reduced inference throughput, and operating system instability under continuous AI workloads. Reliability-oriented deployment strategies for assistive edge-AI systems therefore remain an important open research challenge.
	
	Table~\ref{tab:comparison} summarizes the characteristics and limitations of representative assistive mobility systems reported in prior literature. In contrast to existing approaches, the proposed framework emphasizes the combination of affordability, fully offline edge inference, multimodal environmental awareness, and reliability-oriented embedded deployment on severely constrained hardware.
	
	\begin{table*}[t]
		\caption{Comparison of Representative Assistive Mobility Systems. Cost figures are approximate and sourced from cited literature.}
		\label{tab:comparison}
		\centering
		\small
		\setlength{\tabcolsep}{4pt}
		\renewcommand{\arraystretch}{1.2}
		\resizebox{\textwidth}{!}{
			\begin{tabular}{p{3cm} p{2cm} p{2cm} p{2cm} p{2cm} p{2cm} p{4.5cm}}
				\toprule
				\textbf{System / Work} & \textbf{Primary Modality} & \textbf{Processing Platform} & \textbf{Typical Latency} & \textbf{Offline Operation} & \textbf{Estimated Cost} & \textbf{Primary Limitations} \\
				\midrule
				
				NavBelt~\cite{shoval1998navbelt}
				& Ultrasonic sensing
				& Microcontroller
				& $<$50\,ms
				& Yes
				& $\sim$\$50
				& Limited semantic awareness and difficulty distinguishing obstacle context. \\
				
				Tapu et al.~\cite{tapu2020survey}
				& Wearable camera-based vision
				& High-end smartphone
				& $\sim$200\,ms
				& Partial
				& $\sim$\$800
				& Increased thermal load, ergonomic limitations, and battery drain during continuous operation. \\
				
				Upadhyaya et al.~\cite{upadhyaya2024real}
				& RGB vision (YOLOv8)
				& Raspberry Pi 4
				& $\sim$250\,ms
				& Yes
				& $\sim$\$200
				& Higher power consumption and larger hardware footprint compared to lightweight assistive systems. \\
				
				OrCam MyEye~\cite{Lavric2024}
				& Wearable vision sensing
				& Proprietary embedded hardware
				& $\sim$100\,ms
				& Yes
				& $\sim$\$3{,}500--4{,}500
				& High commercial cost and limited accessibility in low-resource settings. \\
				
				Cloud-Assisted Frameworks~\cite{chen2019edge}
				& RGB + LiDAR
				& Edge-to-cloud infrastructure
				& $>$500\,ms
				& No
				& Variable
				& Dependence on network connectivity and variable communication latency. \\
				
				\midrule
				
				\textbf{Proposed System}
				& \textbf{RGB + ToF fusion}
				& \textbf{Raspberry Pi Zero 2W}
				& \textbf{$\sim$330\,ms}
				& \textbf{Yes}
				& \textbf{$\sim$\$88}
				& \textbf{Reduced frame rate and limited field-of-view under constrained hardware conditions.} \\
				
				\bottomrule
			\end{tabular}
		}
	\end{table*}

	\section{AI-Integrated Smart Cane Architecture}
	\label{sec:arch}
	
\begin{figure}[t]
		\centering
		\includegraphics[width=0.5\textwidth]{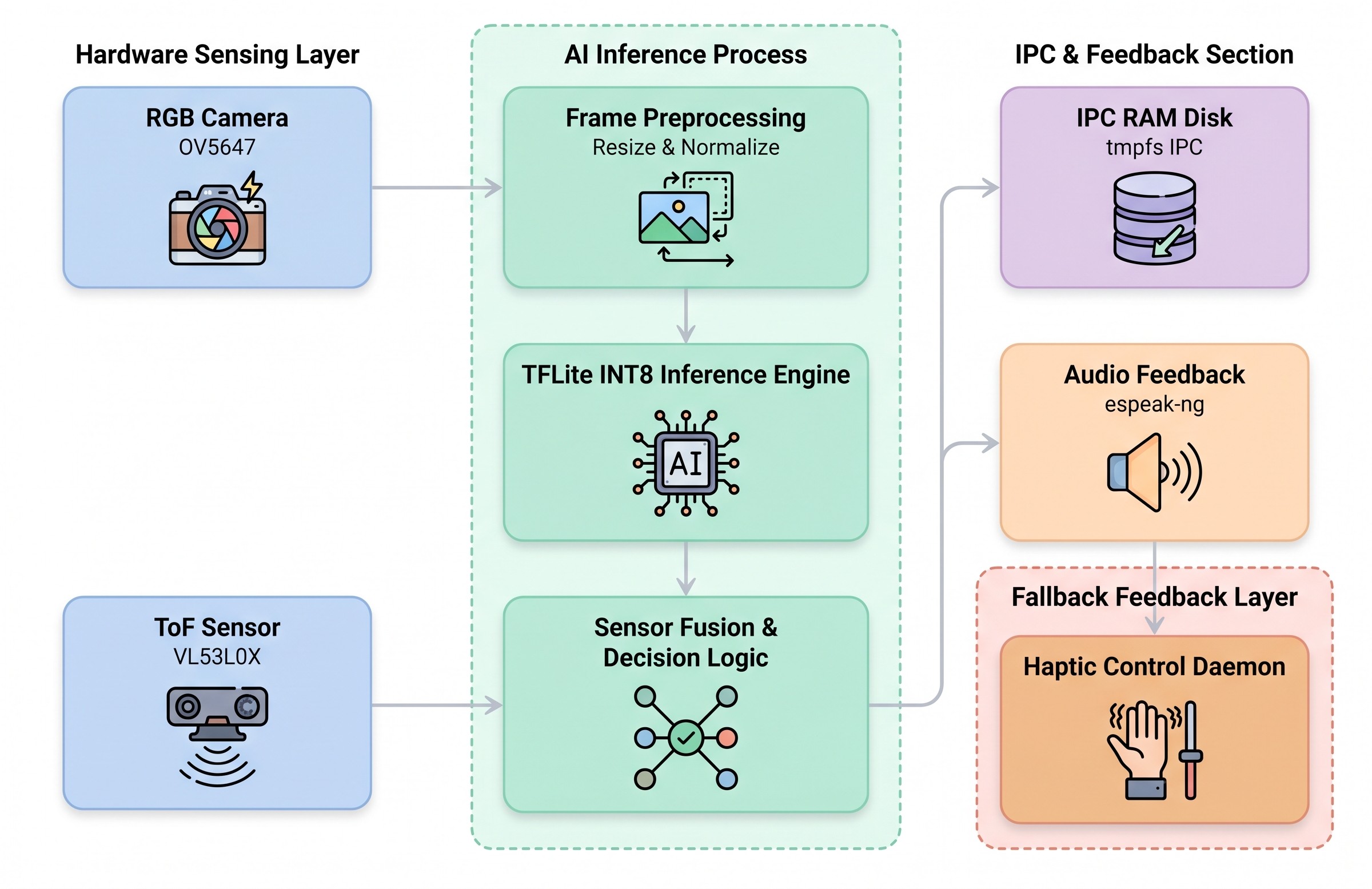}
		\caption{Overview of the proposed AI-integrated smart cane architecture. Sensing, inference, and assistive feedback are separated into isolated processes. Audio feedback (\texttt{espeak-ng}) is triggered directly by the inference process for low-latency verbal notifications, while haptic control operates as an independent daemon communicating via a lightweight \texttt{tmpfs}-based IPC mechanism. This separation enables continued vibrotactile feedback during inference slowdowns or partial subsystem interruptions.}
		\label{fig:arch}
	\end{figure}

	The proposed system is designed as a modular embedded assistive mobility framework intended for reliable operation under severe computational and energy constraints. Unlike monolithic prototype implementations that tightly couple sensing, inference, and feedback generation within a single execution pipeline, the proposed architecture separates perception, decision-making, and assistive feedback into isolated processes to improve responsiveness and operational robustness during extended real-world usage, as illustrated in Fig.~\ref{fig:arch}.
	
	\subsection{Physical Smart Cane Design}
	
	The hardware platform is integrated onto a conventional 120\,cm aluminum mobility cane to preserve familiar user interaction patterns and minimize additional training requirements. The embedded electronics, including the Raspberry Pi Zero 2W, power regulation circuitry, and a dual-cell 5000\,mAh lithium-ion battery pack, are housed within a compact 3D-printed enclosure positioned near the cane grip to maintain ergonomic balance during prolonged usage. The complete prototype weighs approximately 380\,g (excluding the base cane structure).
	
\begin{figure}[t]
		\centering
		\includegraphics[width=0.5\textwidth]{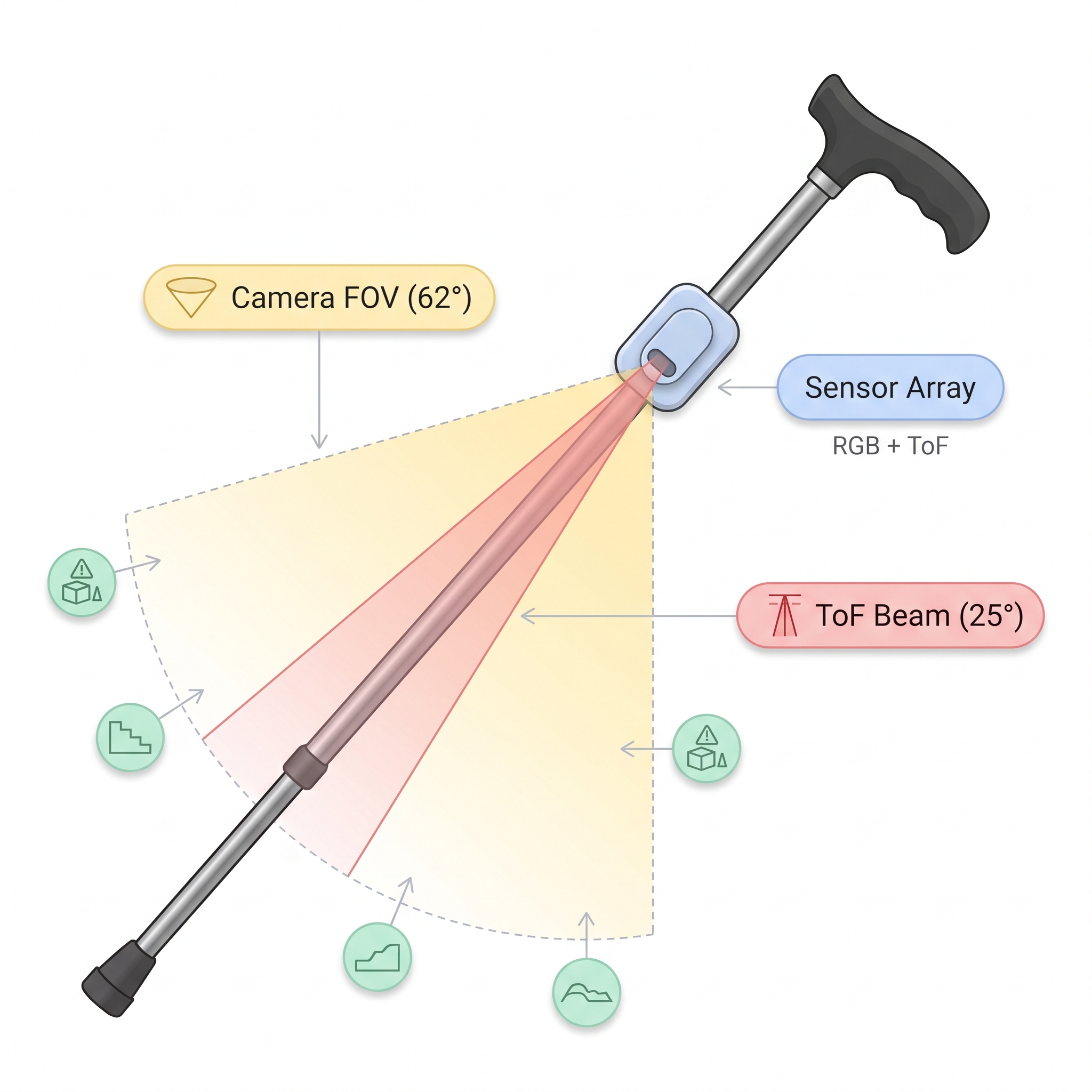}
		\caption{Sensor placement and spatial sensing coverage. The wider RGB field-of-view ($62^\circ$) supports semantic scene awareness, while the narrower ToF beam ($25^\circ$) provides localized depth estimation for nearby obstacle detection.}
		\label{fig:cane_sensor}
	\end{figure}

	Sensor placement was selected to maximize forward environmental awareness while maintaining ergonomic handling characteristics. The RGB camera and ToF sensor are mounted coaxially with an approximate downward tilt angle of $45^\circ$, enabling simultaneous forward obstacle observation and near-ground distance estimation. As illustrated in Fig.~\ref{fig:cane_sensor}, the RGB camera provides a wider field-of-view for semantic scene interpretation, whereas the ToF sensor provides localized depth measurements for short-range obstacle awareness.
	
	\subsection{Multimodal Sensing Framework}
	
	Environmental perception is achieved through the integration of complementary visual and spatial sensing modalities:
	
	\begin{enumerate}
		\item \textbf{Semantic RGB Sensing:} An OV5647 RGB camera captures $320\times240$ image frames used for semantic obstacle recognition and scene interpretation.
		
		\item \textbf{Spatial ToF Sensing:} A VL53L0X Time-of-Flight sensor continuously measures obstacle distance within an operational range of approximately 10--2000\,mm independent of ambient illumination conditions.
	\end{enumerate}
	
	The combination of RGB and ToF sensing enables partial mitigation of individual sensor limitations. The RGB modality provides semantic understanding but remains sensitive to lighting conditions and motion blur, whereas the ToF sensor provides stable short-range depth estimation but lacks semantic awareness. When semantic obstacles are detected outside the narrow ToF sensing cone, the system provides qualitative audio alerts without explicit distance estimation.
	
	\subsection{Embedded AI Pipeline}
	
	Deploying deep learning inference on the Raspberry Pi Zero 2W presents substantial memory and thermal constraints due to the platform's 512\,MB RAM limitation and absence of dedicated AI acceleration hardware. To balance computational feasibility and semantic detection capability, the proposed framework utilizes an SSD MobileNet V1 model~\cite{howard2017mobilenets,liu2016ssd} optimized using TensorFlow Lite INT8 quantization~\cite{jacob2018quantization}. SSD MobileNet V1 was selected over V2~\cite{sandler2018mobilenetv2} and lightweight alternatives such as EfficientDet-Lite~\cite{tan2020efficientdet} primarily due to its lower post-quantization memory footprint (approximately 4.8\,MB for INT8 V1 versus approximately 7.2\,MB for INT8 V2), which is critical given the 512\,MB shared-memory constraint of the target platform. Although MobileNet V2 and V3 offer improved accuracy--efficiency tradeoffs on general benchmarks, the additional memory overhead was found to introduce instability under concurrent multiprocess operation during preliminary testing.
	
	Captured image frames undergo resizing and normalization before inference execution. To reduce sustained CPU load and thermal accumulation during prolonged operation, the system implements a fixed modulo-3 frame skipping policy, in which semantic inference is executed on every third captured frame ($f_{\text{infer}} = f_{\text{capture}} / 3$), resulting in an effective inference throughput of approximately 2--3\,FPS under typical operating conditions. In parallel, the ToF sensor continues independent high-frequency polling at approximately 20\,Hz to maintain short-range obstacle awareness between visual inference cycles.
	
	\subsection{Assistive Feedback Framework}
	
	The proposed system employs multimodal assistive feedback to reduce dependence on any single sensory channel and minimize user cognitive workload during navigation tasks.
	
	\begin{itemize}
		\item \textbf{Audio-Based Guidance:} Semantic obstacle notifications are generated using the lightweight \texttt{espeak-ng} speech synthesis engine and delivered through Bluetooth audio output. Audio prompts are intentionally concise to reduce distraction during mobility tasks.
		
		\item \textbf{Vibrotactile Feedback:} An eccentric rotating mass (ERM) vibration motor provides proximity-aware haptic alerts using varying pulse frequencies and durations corresponding to obstacle distance.
	\end{itemize}
	
	Table~\ref{tab:haptic} presents the formal vibrotactile encoding scheme mapping ToF-measured obstacle distance to haptic pulse parameters. At distances exceeding 150\,cm, the system transitions to audio-only informational alerts to reduce haptic fatigue during extended navigation. The encoding boundaries were derived from preliminary calibration trials targeting a user reaction time budget consistent with the 330\,ms mean end-to-end latency and a conservative walking speed assumption of approximately 1.2\,m/s.
	
	\begin{table}[ht]
		\caption{Vibrotactile Feedback Encoding Scheme Based on Obstacle Proximity Distance}
		\label{tab:haptic}
		\centering
		\renewcommand{\arraystretch}{1.15}
		\resizebox{\columnwidth}{!}{
			\begin{tabular}{lcccc}
				\toprule
				\textbf{Distance Range} & \textbf{Pulse Duration (ms)} & \textbf{Pulse Interval (ms)} & \textbf{Alert Priority} \\
				\midrule
				$<$50\,cm & 500 & 100 & Critical \\
				50--80\,cm & 300 & 200 & High \\
				80--120\,cm & 200 & 400 & Moderate \\
				120--150\,cm & 100 & 800 & Low \\
				$>$150\,cm & Audio only & --- & Informational \\
				\bottomrule
			\end{tabular}
		}
	\end{table}
	
	To reduce the perceptual impact of intermittent inference gaps caused by constrained frame rates, the system maintains a short temporal persistence window for obstacle alerts. This mechanism smooths rapid detection fluctuations and reduces abrupt feedback discontinuities during user movement.
	
	\subsection{Reliability and Safety Layer}
	
	Reliability and execution predictability represent critical challenges for embedded assistive AI systems operating on resource-constrained Linux platforms~\cite{simmonds2017embedded}. Under high CPU utilization, single-threaded inference applications may delay GPIO responsiveness or temporarily interrupt assistive feedback generation. Ensuring dependable execution in such constrained environments draws upon established principles of lightweight runtime monitoring and behavior attestation~\cite{ismail2014design}, where isolating concurrent tasks prevents system stalls and preserves operational integrity.
	
	To improve operational robustness, the proposed framework adopts a lightweight multiprocessing architecture in which sensing, inference, and haptic control operate as partially isolated processes. Semantic inference outputs are exchanged through a volatile \texttt{tmpfs}-based inter-process communication (IPC) mechanism to avoid excessive SD card writes and reduce storage latency. A lightweight haptic control daemon independently monitors proximity sensing and feedback generation, enabling continued obstacle awareness even during temporary inference interruptions or AI subsystem restarts.
	
	Although the proposed architecture does not eliminate all latency-related limitations, the separation of assistive feedback from the primary inference loop improves responsiveness and reduces the likelihood of complete assistive feedback loss during extended embedded operation.

	\section{Embedded AI Optimization}
	\label{sec:optimization}
	
	Deploying deep neural network inference on ultra-low-cost embedded hardware requires careful optimization across memory usage, computational workload, thermal behavior, and power consumption~\cite{technologies14040215}. Due to the limited computational resources of the Raspberry Pi Zero 2W platform, several software- and system-level optimizations were implemented to maintain stable assistive operation during prolonged usage scenarios.
	
	\subsection{Quantization and Memory Optimization}
	
	The SSD MobileNet V1 model was converted from TensorFlow to TensorFlow Lite (TFLite) format and optimized using Post-Training Quantization (PTQ)~\cite{jacob2018quantization}, reducing model weights from 32-bit floating-point precision to 8-bit integer (INT8) representation. Quantization reduced the model storage footprint from approximately 22.5\,MB to 4.8\,MB while also reducing memory bandwidth requirements during inference execution. To quantify the accuracy impact of this compression step, the FP32 and INT8 model variants were evaluated on the same custom indoor encounter dataset. The INT8 model exhibited a macro-averaged F1-score reduction of approximately 1.5 percentage points relative to the FP32 baseline, consistent with quantization degradation ranges reported for SSD MobileNet INT8 conversion in the literature~\cite{jacob2018quantization}, and was considered an acceptable tradeoff given the memory and throughput improvements achieved.
	
	In addition to model quantization, memory usage was further reduced through lightweight frame preprocessing, fixed-resolution image capture, and separation of inference and feedback tasks into independent processes. These optimizations were necessary to avoid memory exhaustion on the Raspberry Pi Zero 2W platform, which provides only 512\,MB of shared system memory without dedicated AI acceleration hardware.
	
	Table~\ref{tab:profiling} summarizes approximate resource utilization measurements observed during continuous system operation. Measurements were collected using Linux system monitoring utilities (\texttt{top}, \texttt{vcgencmd measure\_temp}) under sustained indoor testing conditions. The peak system load row represents concurrent execution of all subsystems during active inference cycles; average power draw between inference frames, when only OS services, camera capture, and ToF polling are active, is estimated at approximately 1.5--1.8\,W.
	
	\begin{table}[ht]
		\caption{Approximate System Resource Profiling on Raspberry Pi Zero 2W During Continuous Operation}
		\label{tab:profiling}
		\centering
		\renewcommand{\arraystretch}{1.15}
		\resizebox{\columnwidth}{!}{
			\begin{tabular}{lccc}
				\toprule
				\textbf{Subsystem} & \textbf{CPU Usage (\%)} & \textbf{RAM (MB)} & \textbf{Power (W)} \\
				\midrule
				
				Background OS Services
				& 7--10
				& 105--120
				& 0.7--0.9 \\
				
				Camera Capture + Preprocessing
				& 12--18
				& 40--50
				& 0.3--0.5 \\
				
				TFLite INT8 Inference
				& 60--70
				& 170--190
				& 1.1--1.3 \\
				
				ToF + Haptic Feedback Daemon
				& 1--3
				& 6--10
				& 0.3--0.4 \\
				
				\midrule
				
				\textbf{Peak Concurrent System Load}
				& \textbf{85--92}
				& \textbf{330--360}
				& \textbf{2.6--2.9} \\
				
				\bottomrule
			\end{tabular}
		}
	\end{table}
	
	Observed profiling results indicate that the quantized inference engine remains the dominant contributor to both CPU utilization and memory consumption. Nevertheless, the optimized pipeline maintained stable operation without repeated out-of-memory (OOM) termination events during extended testing sessions.
	
	\subsection{Inference Scheduling and Latency Optimization}
	
	Sustained real-time semantic inference on the Raspberry Pi Zero 2W is constrained by limited CPU throughput and thermal headroom. To reduce processing overhead while preserving obstacle awareness functionality, the system employs the fixed modulo-3 frame skipping policy described in Section~\ref{sec:arch}, which reduces sustained CPU utilization and thermal accumulation during prolonged operation while allowing the ToF sensing subsystem to continue high-frequency obstacle polling independently.
	
\begin{figure}[t]
		\centering
		\includegraphics[width=0.5\textwidth]{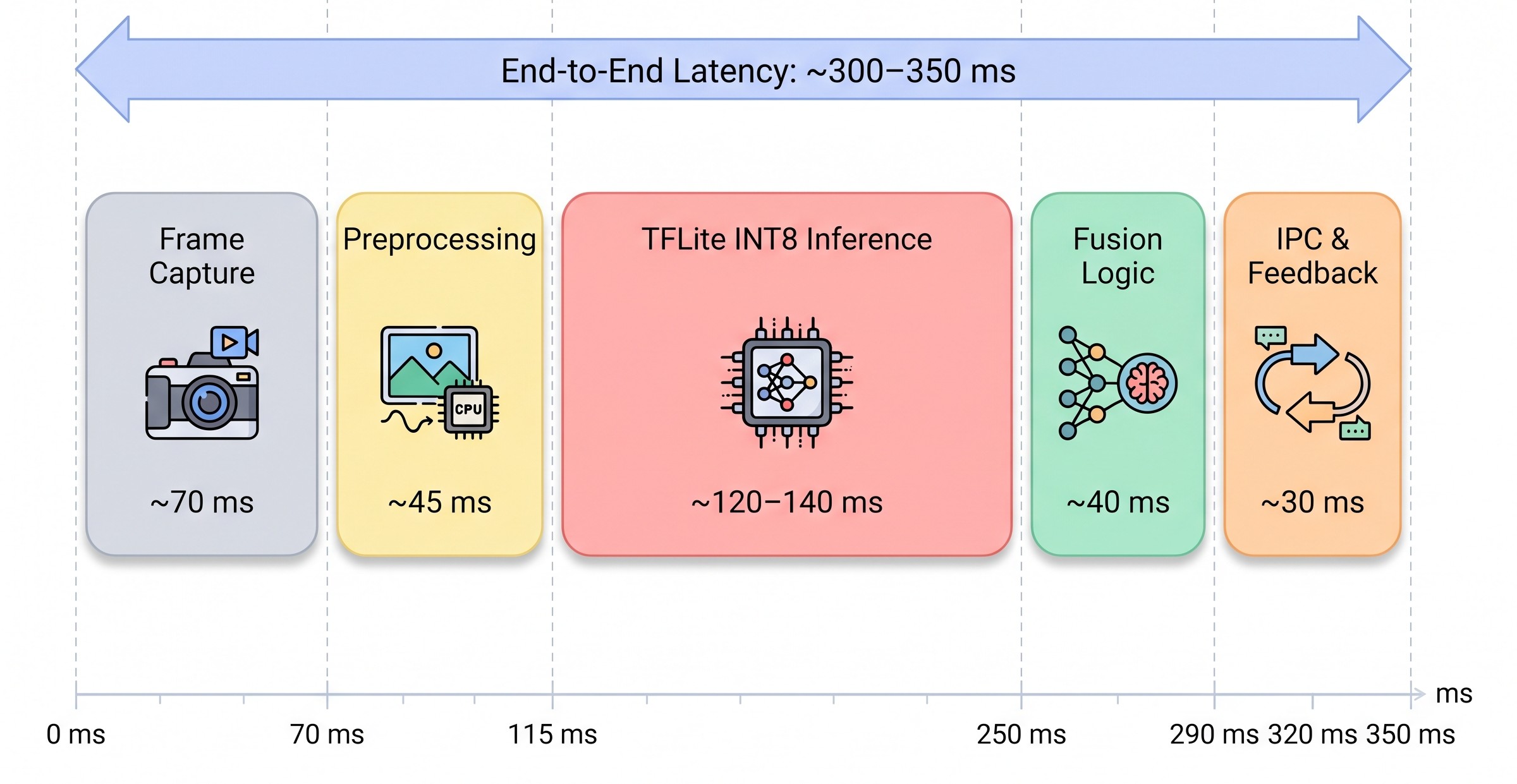}
		\caption{Approximate end-to-end latency breakdown for a representative semantic detection cycle on the Raspberry Pi Zero 2W platform. Stage boundaries and widths are proportional to measured duration on the horizontal time axis (ms). Neural inference remains the dominant computational bottleneck, contributing approximately 120--140\,ms of the total 300--350\,ms pipeline latency.}
		\label{fig:latency}
	\end{figure}

	Figure~\ref{fig:latency} illustrates the approximate end-to-end processing latency breakdown for a representative obstacle detection cycle. The timeline positions each processing stage at its approximate start time on the horizontal axis. The largest temporal contribution originates from TFLite inference execution, followed by frame acquisition and preprocessing overhead.
	
	Although the achieved inference throughput does not support high-speed autonomous navigation, the optimized embedded pipeline provides sufficient responsiveness for supplementary environmental awareness assistance during moderate walking-speed mobility scenarios.

	\section{Human-Centered Assistive Design}
	\label{sec:human_centered}
	
	The proposed system is designed as a supplementary mobility assistance tool intended to enhance environmental awareness rather than replace conventional white cane techniques or formal orientation and mobility training. Users continue to rely on standard cane sweeping mechanics for ground-level tactile interpretation, while the embedded AI layer provides additional awareness regarding elevated, dynamic, or semantically relevant obstacles that may otherwise remain undetected through traditional tactile interaction alone.
	
	A primary design objective was the reduction of cognitive workload during mobility tasks. Continuous environmental narration was intentionally avoided to prevent excessive auditory distraction and alert fatigue. Instead, the system follows an event-driven assistive feedback strategy in which notifications are generated only when potentially relevant environmental changes or obstacles are detected.
	
	Assistive feedback intensity is dynamically adapted according to estimated obstacle proximity and contextual urgency, following the encoding scheme defined in Table~\ref{tab:haptic}. Distant semantic detections ($>150$\,cm) generate low-priority informational alerts, whereas nearby obstacles ($<80$\,cm) trigger increasingly aggressive vibrotactile feedback patterns. Under close-range conditions, haptic warnings are prioritized over verbal notifications to minimize reaction latency and reduce auditory interruption during navigation.
	
	In addition, the multimodal feedback strategy was designed to preserve user situational awareness rather than dominate it. Short-duration speech prompts and lightweight haptic encoding were selected to minimize attentional fragmentation while still conveying meaningful environmental information under constrained embedded sensing conditions.

	\section{Evaluation and Results}
	\label{sec:eval}
	
	The proposed smart cane system was evaluated across technical performance, semantic detection capability, mobility assistance effectiveness, and preliminary usability dimensions. All evaluations were conducted primarily in indoor navigation scenarios under controlled but realistic environmental conditions.
	
	\subsection{Software Environment and Experimental Setup}
	
	All experiments were conducted on the Raspberry Pi Zero 2W running Raspberry Pi OS Lite (64-bit, Debian Bookworm, kernel 6.1), with TensorFlow Lite Runtime 2.13, Python 3.11, the \texttt{picamera2} library for camera interfacing, and \texttt{smbus2} for I$^2$C sensor communication. System resource measurements were obtained using standard Linux monitoring tools (\texttt{top}, \texttt{vcgencmd measure\_temp}). All inference experiments were conducted with the CPU frequency governor set to \texttt{ondemand} to reflect realistic deployment conditions.
	
	\subsection{Custom Evaluation Dataset}
	
	Semantic obstacle detection performance was evaluated using a custom indoor evaluation dataset comprising approximately 600 manually annotated obstacle encounter instances spanning five semantic categories relevant to indoor navigation. The base SSD MobileNet V1 model pretrained on COCO was deployed without fine-tuning on the custom dataset; the custom encounters were used exclusively for detection threshold calibration and performance evaluation. Annotation was performed through manual labeling by two independent raters, with inter-rater disagreements resolved through consensus review. No train/test split was applied to the evaluation dataset, as the pretrained model was not retrained; the full 600 encounters constitute the evaluation set. Class instances were approximately balanced across the five categories, with minor natural overrepresentation of person and chair instances in the cluttered corridor scenario.
	
	\subsection{Technical Performance Evaluation}
	
	End-to-end response latency represents a critical consideration for embedded assistive mobility systems. Based on repeated profiling measurements, the proposed framework achieved an average semantic processing latency of approximately 300--350\,ms per inference cycle under sustained operation on the Raspberry Pi Zero 2W platform. Assuming an average walking speed of approximately 1.1--1.3\,m/s, this latency corresponds to roughly 35--45\,cm of forward user movement before assistive feedback delivery.
	
	To partially compensate for inference delay variability, the ToF sensing subsystem continuously monitors short-range obstacle proximity independently of the semantic inference pipeline. During experimental testing, the fallback haptic warning threshold was configured conservatively at approximately 80\,cm to preserve additional user reaction distance under uncertain detection conditions.
	
	Thermal measurements collected during continuous operation in a controlled indoor environment ($24^\circ$C ambient temperature) using \texttt{vcgencmd measure\_temp} indicated steady-state processor temperatures ranging between approximately $65^\circ$C and $70^\circ$C without active cooling during a two-hour continuous test session. No sustained thermal throttling events were observed during the evaluation period; however, prolonged outdoor operation under elevated ambient temperatures was not investigated in this study and represents an important open question for future characterization.
	
	\subsection{AI Detection Evaluation}
	
	Table~\ref{tab:ai_metrics} summarizes the semantic obstacle detection performance across the five evaluated indoor categories. With approximately 120 encounters evaluated per class (600 total encounters), the false positive (FP) rates per 100 negative instances ranged from 1.7 (person) to 7.5 (potted plant).
	
	\begin{table}[ht]
		\caption{Semantic Obstacle Detection Performance Across Indoor Categories (N$\approx$120 per class, 600 total encounters)}
		\label{tab:ai_metrics}
		\centering
		\renewcommand{\arraystretch}{1.15}
		\resizebox{\columnwidth}{!}{
			\begin{tabular}{lcccc}
				\toprule
				\textbf{Obstacle Class} & \textbf{Precision} & \textbf{Recall} & \textbf{F1-Score} & \textbf{FP per 100 neg.} \\
				\midrule
				
				Person
				& 0.94
				& 0.91
				& 0.92
				& 1.7 \\
				
				Chair
				& 0.88
				& 0.85
				& 0.86
				& 3.3 \\
				
				Desk / Table
				& 0.86
				& 0.82
				& 0.84
				& 4.2 \\
				
				Doorway
				& 0.82
				& 0.79
				& 0.80
				& 5.8 \\
				
				Potted Plant
				& 0.75
				& 0.68
				& 0.71
				& 7.5 \\
				
				\midrule
				
				\textbf{Macro Average}
				& \textbf{0.85}
				& \textbf{0.81}
				& \textbf{0.82}
				& --- \\
				
				\bottomrule
			\end{tabular}
		}
	\end{table}
	
	As summarized in Table~\ref{tab:ai_metrics}, the proposed framework achieved an overall macro F1-score of 0.82 across evaluated semantic classes. Detection performance remained strongest for large, visually distinct obstacle categories such as pedestrians (person class) and chairs. The high person-class precision (0.94) and recall (0.91) can be attributed to the strong person-class representation in the COCO pretrained model and the relatively unambiguous silhouette signature of standing pedestrians under controlled indoor illumination. Performance degradation was observed for geometrically irregular or partially occluded objects including decorative plants and narrow structures. Reduced image resolution and constrained inference throughput also contributed to occasional missed detections under rapid motion conditions.
	
	\subsection{Mobility Assistance Evaluation}
	
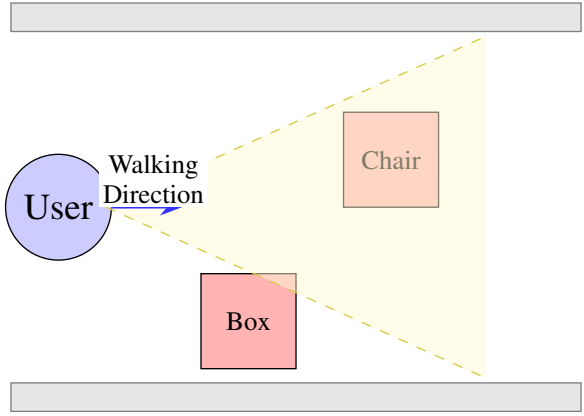
\begin{figure}[t]
		\centering
		\resizebox{0.9\columnwidth}{!}{
			\begin{tikzpicture}[
				user/.style={circle, draw=black, fill=blue!20, minimum size=0.8cm},
				obs/.style={rectangle, draw=black, fill=red!30, minimum width=1cm, minimum height=1cm},
				wall/.style={rectangle, draw=gray, fill=gray!20, minimum width=6cm, minimum height=0.3cm},
				fov/.style={draw=none, fill=yellow!20, opacity=0.5}
				]
				
				\node[wall] (w1) at (3,3) {};
				\node[wall] (w2) at (3,-1) {};
				
				\node[user] (u) at (0.5,1) {User};
				\node[obs, align=center, font=\scriptsize] (o1) at (4,1.5) {Chair};
				\node[obs, align=center, font=\scriptsize] (o2) at (2.5,-0.2) {Box};
				
				\fill[fov] (1.0,1) -- (5,2.8) -- (5,-0.8) -- cycle;
				\draw[dashed, draw=yellow!80!black] (1.0,1) -- (5,2.8);
				\draw[dashed, draw=yellow!80!black] (1.0,1) -- (5,-0.8);
				
				\draw[-{Stealth}, thick, draw=blue!80] (u) -- (1.8,1);
				
				\node[font=\scriptsize, align=center, fill=white, inner sep=1pt] at (1.5,1.3) {Walking\\Direction};
				
			\end{tikzpicture}
		}
		\caption{Representative indoor corridor evaluation scenario illustrating semantic and spatial obstacle monitoring along the user walking path. The shaded cone represents the approximate RGB camera field-of-view during forward navigation.}
		\label{fig:mobility}
	\end{figure}

	To evaluate practical assistive functionality, the proposed system was tested across multiple indoor mobility scenarios including cluttered corridor traversal, elevated obstacle avoidance, and dynamic pedestrian encounters. Figure~\ref{fig:mobility} illustrates the representative corridor evaluation environment used during testing.
	
	Table~\ref{tab:mobility} reports obstacle avoidance success rates with 95\% confidence intervals computed using the Wilson score method, along with mean reaction times for each evaluated scenario. With N=30 binary trials per scenario, the Wilson 95\% confidence intervals are approximately $\pm$10--14 percentage points, reflecting the inherent statistical uncertainty at this sample size; these intervals are reported transparently to enable appropriate interpretation of the point estimates.
	
	\begin{table}[ht]
		\caption{Mobility Assistance Evaluation Results (N=30 trials per scenario; 95\% CI computed using Wilson score method)}
		\label{tab:mobility}
		\centering
		\renewcommand{\arraystretch}{1.15}
		\resizebox{\columnwidth}{!}{
			\begin{tabular}{lcc}
				\toprule
				\textbf{Scenario} & \textbf{Avoidance Rate (95\% CI)} & \textbf{Mean Reaction Time} \\
				\midrule
				
				Cluttered Corridor
				& 93.3\% [78.7\%, 98.1\%] (28/30)
				& 1.1\,s \\
				
				Elevated Hazard
				& 86.7\% [70.3\%, 94.7\%] (26/30)
				& 1.4\,s \\
				
				Dynamic Pedestrian
				& 80.0\% [62.7\%, 90.5\%] (24/30)
				& 1.8\,s \\
				
				\bottomrule
			\end{tabular}
		}
	\end{table}
	
	The proposed framework demonstrated the strongest performance during structured corridor navigation tasks, while dynamic obstacle encounters produced increased reaction times and occasional delayed semantic recognition events. Elevated obstacle scenarios showed improved detection capability relative to conventional white cane interaction, although failures occasionally occurred during rapid cane sweeping motions or when narrow obstacles briefly exited the ToF sensing region.
	
	\subsection{Preliminary User Study and Usability Assessment}
	
	A preliminary usability assessment was conducted with 12 participants under simulated visual impairment conditions using blindfold-assisted navigation tasks within a controlled indoor obstacle course. Participants completed short-range mobility tasks while using the proposed smart cane system under supervised safety conditions.
	
	Usability and perceived workload were assessed using the System Usability Scale (SUS)~\cite{brooke1996sus} and the NASA Task Load Index (NASA-TLX)~\cite{hart1988nasa}, respectively, with results summarized in Table~\ref{tab:usability}. The SUS is scored on a 0--100 scale, with scores above 68 considered above-average usability.
	
	\begin{table}[ht]
		\caption{Preliminary Usability Assessment Results (N=12 blindfold-assisted participants)}
		\label{tab:usability}
		\centering
		\renewcommand{\arraystretch}{1.15}
		\resizebox{\columnwidth}{!}{
			\begin{tabular}{lcc}
				\toprule
				\textbf{Metric} & \textbf{Mean Score (95\% CI)} & \textbf{Interpretation} \\
				\midrule
				
				SUS Score
				& 78.5 ($\pm$4.2) [71.0, 86.0]
				& Above average usability (threshold: 68) \\
				
				NASA-TLX Mental Demand
				& 45 / 100
				& Moderate workload \\
				
				NASA-TLX Frustration
				& 22 / 100
				& Relatively low frustration \\
				
				Perceived Safety
				& 4.2 / 5.0
				& Positive user perception \\
				
				\bottomrule
			\end{tabular}
		}
	\end{table}
	
	The mean SUS score of 78.5 exceeds the widely adopted 68-point threshold for above-average usability~\cite{brooke1996sus}, suggesting that participants found the system usable within the constrained evaluation context. Participants generally reported that the combined haptic and audio feedback improved environmental awareness and increased confidence during obstacle navigation tasks. Several participants specifically identified semantic obstacle labeling as beneficial in cluttered indoor scenarios. However, users also reported occasional delayed feedback during rapid movement and noted limitations associated with the restricted field-of-view of the sensing hardware.
	
	The preliminary usability study remains limited in scale and participant composition. The use of blindfolded sighted participants rather than experienced visually impaired white cane users is an acknowledged methodological limitation; SUS norms derived from this population may not generalize to long-term experienced blind users. Accordingly, the preliminary findings are reported as indicative pilot data rather than definitive usability conclusions, and broader validation with experienced visually impaired participants remains an essential direction for future work.

	\section{Failure Analysis and Limitations}
	\label{sec:failures}
	
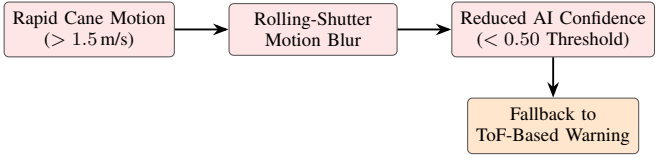
\begin{figure}[t]
		\centering
		\resizebox{\columnwidth}{!}{
			\begin{tikzpicture}[
				box/.style={rectangle, draw=black!70, fill=red!10, minimum width=2.5cm, align=center, font=\scriptsize, rounded corners=2pt},
				ar/.style={-{Stealth}, thick}
				]
				
				\node[box] (sweep) {Rapid Cane Motion\\($>1.5$\,m/s)};
				
				\node[box, right=0.8cm of sweep] (blur) {Rolling-Shutter\\Motion Blur};
				
				\node[box, right=0.8cm of blur] (conf) {Reduced AI Confidence\\($<0.50$ Threshold)};
				
				\node[box, fill=orange!20, below=0.6cm of conf] (fail) {Fallback to\\ToF-Based Warning};
				
				\draw[ar] (sweep) -- (blur);
				\draw[ar] (blur) -- (conf);
				\draw[ar] (conf) -- (fail);
				
			\end{tikzpicture}
		}
		\caption{Illustrative failure pathway associated with rapid cane motion. Motion blur introduced by the rolling-shutter camera reduces semantic inference confidence below the 0.50 acceptance threshold, increasing dependence on short-range ToF-based obstacle awareness as a fallback mechanism.}
		\label{fig:failure}
	\end{figure}

	Despite demonstrating promising assistive functionality under controlled indoor conditions, the proposed edge-AI smart cane remains subject to several important sensing, computational, and deployment limitations that constrain real-world performance. Transparent discussion of these limitations is essential for responsible evaluation of embedded assistive mobility systems, including the primary failure pathways induced by rapid cane movement (Fig.~\ref{fig:failure}).
	
	Table~\ref{tab:failures} summarizes representative failure modes and operational constraints observed during experimental evaluation. One of the most significant limitations arises from rapid cane sweeping motion, which introduces motion blur artifacts due to the rolling-shutter characteristics of the low-cost RGB camera module. Under these conditions, semantic inference confidence may degrade substantially, particularly for small or thin obstacles.
	
	\begin{table}[ht]
		\caption{Observed Failure Modes and Deployment Limitations}
		\label{tab:failures}
		\centering
		\renewcommand{\arraystretch}{1.15}
		\begin{tabularx}{\columnwidth}{>{\raggedright\arraybackslash}p{2.3cm} >{\raggedright\arraybackslash}X}
			\toprule
			\textbf{Failure Scenario} & \textbf{Observed Impact and Current Mitigation} \\
			\midrule
			
			Rapid Cane Motion
			& Fast sweeping motion introduces motion blur and intermittent semantic detection instability. Partial mitigation is provided through independent ToF-based proximity sensing. \\
			
			Low-Light Environments
			& RGB semantic inference performance degrades substantially under poor illumination conditions. Under these scenarios, the system relies primarily on short-range ToF sensing. \\
			
			Reflective or Glass Surfaces
			& Specular reflections may distort ToF measurements and produce unstable distance estimates. This limitation remains only partially addressed in the current implementation. \\
			
			Narrow Field-of-View
			& Obstacles outside the combined RGB-ToF sensing region may not be detected consistently during rapid directional changes. \\
			
			Embedded Processing Constraints
			& Limited inference throughput ($\sim$2--3 FPS) may introduce delayed semantic feedback during fast user movement or dynamic obstacle encounters. \\
			
			\bottomrule
		\end{tabularx}
	\end{table}
	
	In addition to sensing limitations, the constrained computational capacity of the Raspberry Pi Zero 2W imposes important tradeoffs between inference accuracy, latency, thermal stability, and battery life. Although the multiprocessing architecture improves operational robustness, it does not eliminate all sources of latency variability or intermittent semantic detection failure under challenging environmental conditions.
	
	Furthermore, the current evaluation was limited primarily to controlled indoor scenarios and preliminary blindfold-assisted user studies. Broader validation involving experienced visually impaired users, outdoor navigation conditions, crowded environments, and long-term deployment studies remains necessary before practical real-world adoption can be fully assessed. The dynamic pedestrian scenario failure rate of 20\% (6/30 trials, 95\% CI [9.5\%, 37.3\%]) underscores the need for formal safety analysis in future iterations of the system. In safety-critical assistive edge IoT deployments, adhering to comprehensive ethical AI governance and trustworthy cybersecurity principles~\cite{jan2026eagf} is essential to ensure user physical safety, protect personal sensor telemetry against tampering or spoofing, and maintain dependable fail-safe operation in dynamic public spaces.
	
	\subsection{Future Safety and System Enhancements}
	
	Several improvements may enhance the robustness and usability of future system iterations. First, integrating an inertial measurement unit (IMU) could enable adaptive exposure control or motion-aware inference scheduling to partially mitigate motion blur during rapid cane movement. Second, replacing the eccentric rotating mass (ERM) motor with a Linear Resonant Actuator (LRA) may improve haptic response precision and enable richer tactile encoding patterns for different obstacle categories or urgency levels.
	
	Future work may also investigate lightweight transformer-based vision architectures such as MobileViT~\cite{mobilevit2022}, hybrid convolutional and vision transformer designs for intelligent sensing systems~\cite{shaikh2024advancing}, or recently proposed efficient detector families including EfficientDet-Lite~\cite{tan2020efficientdet} and YOLOv3-based variants~\cite{redmon2018yolov3} to assess their feasibility under the same platform constraints. Additional sensing modalities, including event-based cameras~\cite{eventcamera2023} or ultra-wideband localization, could further improve obstacle awareness in complex and rapidly changing environments. Looking beyond standalone device operation, integrating edge-native assistive devices into broader smart city infrastructure and digital twin ecosystems~\cite{syed2026climate,syed2026agenticdt} represents a compelling avenue for contextual navigation. In such cyber-physical urban frameworks, municipal digital twins can supply macro-scale spatial geometry, infrastructural state updates, and dynamic hazard notifications, while decentralized sensor telemetry from assistive smart canes can contribute real-time accessibility data to urban mobility models. A model ablation study formally comparing RGB-only, ToF-only, and fused pipeline performance under matched evaluation conditions remains an important open experimental contribution.
	
	Finally, larger-scale human-centered evaluation involving visually impaired participants ($N \geq 20$) and outdoor navigation studies will be essential to better understand long-term usability, trust, cognitive workload, and assistive effectiveness under realistic mobility conditions.

	\section{Conclusion}
	\label{sec:conclusion}
	
	This paper presented an affordable AI-integrated smart cane framework designed to provide supplementary environmental awareness assistance for visually impaired mobility scenarios using ultra-low-cost embedded hardware. By combining RGB-based semantic perception with ToF-based spatial sensing, the proposed system demonstrates that multimodal assistive intelligence can be deployed on resource-constrained edge platforms without reliance on cloud connectivity or high-end computing hardware.
	
	The proposed architecture integrates quantized TensorFlow Lite inference, asynchronous multiprocessing, and multimodal assistive feedback to balance semantic awareness, computational feasibility, and operational robustness on the Raspberry Pi Zero 2W platform. Experimental evaluation demonstrated promising obstacle awareness capability across indoor mobility scenarios, achieving a macro-averaged F1-score of 0.82 and peak system power consumption of approximately 2.8\,W, while highlighting important deployment tradeoffs related to latency, sensing limitations, and constrained inference throughput.
	
	Although the current system does not replace traditional mobility aids or professional orientation training, the presented results suggest that low-cost embedded assistive AI systems may provide meaningful supplementary environmental awareness for visually impaired users in structured navigation settings. More broadly, this work contributes toward the development of accessible, privacy-preserving, and resource-efficient assistive intelligence frameworks for mobility support in cost-sensitive and infrastructure-limited environments.

	\bibliographystyle{IEEEtran}
	\bibliography{references}

@misc{who2023,
	author       = {{World Health Organization}},
	title        = {Blindness and vision impairment},
	year         = {2023},
	url = {https://www.who.int/news-room/fact-sheets/detail/blindness-and-visual-impairment}
}

@article{loomis2001navigation,
	author    = {Loomis, Jack M. and Klatzky, Roberta L. and Golledge, Reginald G.},
	title     = {Navigating without vision: Basic and applied research},
	journal   = {Optometry and Vision Science},
	volume    = {78},
	number    = {5},
	pages     = {282--289},
	year      = {2001},
	doi       = {10.1097/00006324-200105000-00011}
}

@article{shoval1998navbelt,
	author    = {Shoval, Shlomo and Borenstein, Johann and Koren, Yoram},
	title     = {The {NavBelt}---{A} computerized travel aid for the blind based on mobile robotics technology},
	journal   = {IEEE Transactions on Biomedical Engineering},
	volume    = {45},
	number    = {11},
	pages     = {1376--1386},
	year      = {1998},
	doi       = {10.1109/10.725334}
}

@article{tapu2020survey,
	author    = {Tapu, Ruxandra and Mocanu, Bogdan and Zaharia, Titus},
	title     = {Wearable assistive devices for visually impaired: A state of the art survey},
	journal   = {Pattern Recognition Letters},
	volume    = {137},
	pages     = {37--52},
	year      = {2020},
	url       = {https://hal.science/hal-01993932v1/document}
}

@article{chen2019edge,
	author = {Chen, Jiasi and Ran, Xukan},
	title = {Deep Learning With Edge Computing: A Review},
	journal = {Proceedings of the IEEE},
	volume = {107},
	number = {8},
	pages = {1655--1674},
	year = {2019},
	doi = {10.1109/JPROC.2019.2921977}
}

@article{howard2017mobilenets,
	author = {Howard, Andrew G. and Zhu, Menglong and Chen, Bo and others},
	title = {{MobileNets}: Efficient Convolutional Neural Networks for Mobile Vision Applications},
	journal = {arXiv preprint arXiv:1704.04861},
	year = {2017},
	doi = {10.48550/arXiv.1704.04861}
}

@inproceedings{liu2016ssd,
	author = {Liu, Wei and Anguelov, Dragomir and Erhan, Dumitru and others},
	title = {{SSD}: Single Shot {MultiBox} Detector},
	booktitle = {European Conference on Computer Vision (ECCV)},
	pages = {21--37},
	year = {2016},
	doi = {10.1007/978-3-319-46448-0_2}
}

@inproceedings{sandler2018mobilenetv2,
	author = {Sandler, Mark and Howard, Andrew and Zhu, Menglong and others},
	title = {{MobileNetV2}: Inverted Residuals and Linear Bottlenecks},
	booktitle = {Proceedings of the IEEE Conference on Computer Vision and Pattern Recognition (CVPR)},
	pages = {4510--4520},
	year = {2018},
	doi = {10.1109/CVPR.2018.00474}
}

@article{jacob2018quantization,
	author    = {Jacob, Benoit and Kligys, Skirmantas and Chen, Bo and Zhu, Menglong and Tang, Matthew and Howard, Andrew and Adam, Hartwig and Kalenichenko, Dmitry},
	title     = {Quantization and Training of Neural Networks for Efficient Integer-Arithmetic-Only Inference},
	journal   = {Proceedings of the IEEE Conference on Computer Vision and Pattern Recognition (CVPR)},
	pages     = {2704--2713},
	year      = {2018},
	doi       = {10.1109/CVPR.2018.00286}
}

@book{simmonds2017embedded,
	author    = {Simmonds, Chris},
	title     = {Mastering Embedded Linux Programming},
	publisher = {Packt Publishing},
	edition   = {2},
	year      = {2017},
	url={https://www.beagleboard.org/books/mastering-embedded-linux-programming}
}

@incollection{brooke1996sus,
	author    = {Brooke, John},
	title     = {{SUS}: A Quick and Dirty Usability Scale},
	booktitle = {Usability Evaluation in Industry},
	publisher = {Taylor \& Francis},
	pages     = {189--194},
	year      = {1996},
	url={https://digital.ahrq.gov/sites/default/files/docs/survey/systemusabilityscale%2528sus%2529_comp%255B1%255D.pdf}
}

@incollection{hart1988nasa,
	author    = {Hart, Sandra G. and Staveland, Lowell E.},
	title     = {Development of {NASA-TLX} (Task Load Index): Results of Empirical and Theoretical Research},
	booktitle = {Human Mental Workload},
	series    = {Advances in Psychology},
	editor    = {Hancock, P. A. and Meshkati, N.},
	publisher = {North-Holland},
	volume    = {52},
	pages     = {139--183},
	year      = {1988},
	doi       = {10.1016/S0166-4115(08)62386-9}
}

@inproceedings{upadhyaya2024real,
	title={Real-time obstacle detection using {YOLOv8} on {Raspberry Pi 4} for visually challenged people},
	author={Upadhyaya, Bijoy Kumar and Pramanik, Pijush Kanti Dutta and Roy, Priyanka and Sen, Rituparna},
	booktitle={International Conference on Smart Computing and Communication},
	pages={221--235},
	year={2024},
	organization={Springer},
	url={https://doi.org/10.1007/978-981-97-1320-2_19}
}

@inproceedings{mobilevit2022,
	author    = {Mehta, Sachin and Rastegari, Mohammad},
	title     = {{MobileViT}: Light-Weight, General-Purpose, and Mobile-Friendly Vision Transformer},
	booktitle = {International Conference on Learning Representations (ICLR)},
	year      = {2022},
	doi       = {10.48550/arXiv.2110.02178}
}

@inproceedings{tan2020efficientdet,
	author    = {Tan, Mingxing and Pang, Ruoming and Le, Quoc V.},
	title     = {{EfficientDet}: Scalable and Efficient Object Detection},
	booktitle = {Proceedings of the IEEE/CVF Conference on Computer Vision and Pattern Recognition (CVPR)},
	pages     = {10781--10790},
	year      = {2020},
	doi       = {10.1109/CVPR42600.2020.01079}
}

@article{redmon2018yolov3,
	author    = {Redmon, Joseph and Farhadi, Ali},
	title     = {{YOLOv3}: An Incremental Improvement},
	journal   = {arXiv preprint arXiv:1804.02767},
	year      = {2018},
	doi       = {10.48550/arXiv.1804.02767}
}

@article{Lavric2024,
	AUTHOR = {Lavric, Alexandru and Beguni, Cătălin and Zadobrischi, Eduard and Căilean, Alin-Mihai and Avătămăniței, Sebastian-Andrei},
	TITLE = {A Comprehensive Survey on Emerging Assistive Technologies for Visually Impaired Persons: Lighting the Path with Visible Light Communications and Artificial Intelligence Innovations},
	JOURNAL = {Sensors},
	VOLUME = {24},
	YEAR = {2024},
	NUMBER = {15},
	ARTICLE-NUMBER = {4834},
	URL = {https://www.mdpi.com/1424-8220/24/15/4834},
	PubMedID = {39123881},
	ISSN = {1424-8220},
	DOI = {10.3390/s24154834}
}

@article{messaoudi2022review,
	author    = {Messaoudi, Mohamed Dhiaeddine and Menelas, Bob-Antoine J. and Mcheick, Hamid},
	title     = {Review of Navigation Assistive Tools and Technologies for the Visually Impaired},
	journal   = {Sensors},
	volume    = {22},
	number    = {20},
	pages     = {7888},
	year      = {2022},
	doi       = {10.3390/s22207888}
}

@article{ABIDI2024,
	title = {A comprehensive review of navigation systems for visually impaired individuals},
	journal = {Heliyon},
	volume = {10},
	number = {11},
	pages = {e31825},
	year = {2024},
	issn = {2405-8440},
	doi = {https://doi.org/10.1016/j.heliyon.2024.e31825},
	url = {https://www.sciencedirect.com/science/article/pii/S2405844024078563},
	author = {Mustufa Haider Abidi and Arshad {Noor Siddiquee} and Hisham Alkhalefah and Vishwaraj Srivastava}
}

@inproceedings{Nikanfar2025,
	author = {Nikanfar, Sama and Hebri, Aref and Ram Nambiappan, Harish and Nale, Gaurav and Siddiqua, Mahfuza and Farhanipad, Farnaz and Makedon, Fillia},
	title = {A Survey on Assistive Technologies for Visually Impaired Individuals: Recent Innovations, Limitations, and Future Directions},
	year = {2025},
	isbn = {9798400714023},
	publisher = {Association for Computing Machinery},
	address = {New York, NY, USA},
	url = {https://doi.org/10.1145/3733155.3734895},
	doi = {10.1145/3733155.3734895},
	booktitle = {Proceedings of the 18th ACM International Conference on PErvasive Technologies Related to Assistive Environments},
	pages = {429–434},
	numpages = {6}
}

@article{technologies14040215,
	AUTHOR = {Echchidmi, Mohamed and Bouayad, Anas},
	TITLE = {{TinyML} for Sustainable Edge Intelligence: Practical Optimization Under Extreme Resource Constraints},
	JOURNAL = {Technologies},
	VOLUME = {14},
	YEAR = {2026},
	NUMBER = {4},
	ARTICLE-NUMBER = {215},
	URL = {https://www.mdpi.com/2227-7080/14/4/215},
	ISSN = {2227-7080},
	DOI = {10.3390/technologies14040215}
}

@article{eventcamera2023,
	author    = {Gallego, Guillermo and Delbr{\"u}ck, Tobi and Orchard, Garrick and Bartolozzi, Chiara and Taba, Brian and Censi, Andrea and Leutenegger, Stefan and Davison, Andrew J. and Conradt, J{\"o}rg and Daniilidis, Kostas and Scaramuzza, Davide},
	title     = {Event-based vision: A survey},
	journal   = {IEEE Transactions on Pattern Analysis and Machine Intelligence},
	volume    = {44},
	number    = {1},
	pages     = {154--180},
	year      = {2022},
	doi       = {10.1109/TPAMI.2020.3008413}
}

@article{shaikh2024advancing,
  title={Advancing {DDoS} attack detection with hybrid deep learning: integrating convolutional neural networks, {PCA}, and vision transformers},
  author={Shaikh, Jahangir and Syed, Toqeer Ali and Shah, Syed Aziz and Jan, Salman and Ain, Qurat Ul and Singh, Pradeep Kumar},
  journal={International journal on smart sensing and intelligent systems},
  volume={17},
  number={1},
  year={2024},
  publisher={Macquarie University, Australia},
  doi={10.2478/ijssis-2024-0040},
  url={https://doi.org/10.2478/ijssis-2024-0040}
}

@inproceedings{ismail2014design,
author = {Ismail, Roslan and Syed, Toqeer Ali and Musa, Shahrulniza},
title = {Design and implementation of an efficient framework for behaviour attestation using $n$-call slides},
year = {2014},
isbn = {9781450326445},
publisher = {Association for Computing Machinery},
address = {New York, NY, USA},
url = {https://doi.org/10.1145/2557977.2558002},
doi = {10.1145/2557977.2558002},
booktitle = {Proceedings of the 8th International Conference on Ubiquitous Information Management and Communication},
articleno = {36},
numpages = {8},
location = {Siem Reap, Cambodia},
series = {ICUIMC '14}
}

@article{Siddiqui2026ADAPT,
  author  = {Siddiqui, Muhammad Shoaib and Syed, Toqeer Ali and Akarma, Ali},
  title   = {{ADAPT}: An Agentic {AI} Framework for People with Disabilities and Neurodivergence},
  journal = {Journal of Disability Research},
  year    = {2026},
  volume  = {5},
  number  = {2},
  pages   = {e20260830},
  doi     = {10.57197/JDR-2026-0830}
}

@article{syed2026fedagent,
AUTHOR = {Syed, Toqeer Ali and Siddiqui, Muhammad Shoaib and Akarma, Ali and Formisano, Antonio},
TITLE = {{FedAgent-Chain}: A Secure Federated and Agentic {AI} Framework for Multilingual Disability-Inclusive Employment in {AI} Cities},
JOURNAL = {Smart Cities},
VOLUME = {9},
YEAR = {2026},
NUMBER = {7},
ARTICLE-NUMBER = {106},
URL = {https://www.mdpi.com/2624-6511/9/7/106},
ISSN = {2624-6511},
DOI = {10.3390/smartcities9070106}
}

@article{syed2026climate,
AUTHOR = {Syed, Toqeer Ali and Akarma, Ali and Naqash, Muhammad Tayyab and Hameed, Danial and Kamal, Shahid and Formisano, Antonio},
TITLE = {Agentic {AI} for Climate-Resilient Cities: A {PRISMA}-Guided Review and Digital Twin Framework},
JOURNAL = {Sustainability},
VOLUME = {18},
YEAR = {2026},
NUMBER = {17},
ARTICLE-NUMBER = {8917},
URL = {https://www.mdpi.com/2071-1050/18/17/8917},
ISSN = {2071-1050},
DOI = {10.3390/su18178917}
}

@article{syed2026agenticdt,
  title={Agentic {AI}-enhanced digital twins for Smart City civil infrastructure: A secure, autonomous and auditable management framework},
  author={Syed, Toqeer Ali and Akarma, Ali and Alatify, Ali and Naqash, Muhammad Tayyab and Alqurashi, Abdulaziz},
  journal={PLoS One},
  volume={21},
  number={7},
  pages={e0353610},
  year={2026},
  publisher={Public Library of Science},
  doi={10.1371/journal.pone.0353610}
}

@article{jan2026eagf,
  title     = {{EAGF}: A Four-Pillar Ethical {AI} Governance Framework for Trustworthy Cybersecurity in {5G} Renewable Energy {IoT} Systems},
  author    = {Jan, Salman and Akarma, Ali and Syed, Toqeer Ali and Muhammad, Munir Azam and Kamal, Shahid},
  journal   = {Scientific Reports},
  year      = {2026},
  publisher = {Springer Nature},
  doi       = {10.1038/s41598-026-63383-5}
}
	
\end{document}